# Single Image Dehazing through Improved Atmospheric Light Estimation

Huimin Lu[a,b,c,*], Yujie Li[a,*], Shota Nakashima[d,*], Seiichi Serikawa[a,*]

[a] Kyushu Institute of Technology, Japan
[b] Chinese Academy of Sciences, China
[c] Shanghai Jiaotong University, China
[d] Yamaguchi University, Japan

[*]luhuimin@ieee.org

Abstract

Image contrast enhancement for outdoor vision is important for smart car auxiliary transport systems. The video frames captured in poor weather conditions are often characterized by poor visibility. Most image dehazing algorithms consider to use a hard threshold assumptions or user input to estimate atmospheric light. However, the brightest pixels sometimes are objects such as car lights or streetlights, especially for smart car auxiliary transport systems. Simply using a hard threshold may cause a wrong estimation. In this paper, we propose a single optimized image dehazing method that estimates atmospheric light efficiently and removes haze through the estimation of a semi-globally adaptive filter. The enhanced images are characterized with little noise and good exposure in dark regions. The textures and edges of the `processed` images are also enhanced significantly.

1. Introduction

In outdoor imaging, captured images are easily affected by particles in the atmospheric that absorb and scatter light as it travels from the scene to the camera. Such degraded photographs often lack visual vividness, and provide poor visibility of the contents of the scene. This effect may be uncomfortable during driving, as well as for commercial and artistic photographers. Consequently, these real scene must be easily distinguished from the captured images in harsh environments [1].

Research in this field has long focused on image enhancement and dehazing. Schettini et al. [2]

reviewed methods for image enhancement and restoration, as well as subjective and quantitative assessment indices. Integrated color models were proposed for hyper-spectral correction in [3]. Fattal [10] estimated transmission from the albedo of a scene under the assumption of separating object shading and medium transmission. He et al. [11] proposed a dark channel priors to estimate the fog-free outdoor images. Tan [23] used the maximizing local contrast method to recover the scene. These methods typically involve multi-step approaches that use depth information for dehazing. Most image dehazing methods only consider to use a hard threshold assumptions or user input to estimate atmospheric light. Artificial lighting is hardly considered in recent dehazing methods. However, the brightest pixels sometimes are the objects such as car lights or streetlights. The wrong atmospheric light estimation may affects the dehazing results.

In this paper, we propose a single optimized image dehazing method that estimates atmospheric light efficiently and removes haze through the estimation of a semi-globally adaptive filter. The organization of the remainder of this paper is as follows. In Section 2, we review some recent dehazing methods. In Section 3, we propose the light propagation model. Section 4 presents the details of the proposed method. Experimental results are given in Section 5. Finally, we conclude this paper in Section 6.

2. Related Work

Many image dehazing models have been proposed over the past two decades. As per recent haze removal researches, image dehazing can be deconstructed into four main categories:

1) **Polarization**: Schechner et al. [4] restored significantly varied scene distance-images by using a polarization filter attached to a camera. A transmission map of the scene is obtained by capturing two polarized images from different angles. Liang et al. [5] proposed a Stokes matrix-based, four-angle rotation polarization filter to remove haze. However, these methods cannot calculate the transmission of radiance through polarization filters or multiple exposes, especially for time variations that are adverse to visibility and illumination conditions.

2) **Multi-Sensor Fusion**: Narasimhan et al. [6] and Cozman et al. [7] analyzed a static scene by obtaining multiple images under different visibility conditions. Although they reported impressive results, a static camera and a significant change in media turbidity were required under the constant illumination conditions.

3) **HDR Imaging**: Ancuti et al. [9] applied a Laplacian pyramid fusion-based method that considers the pyramid contrast, contrast, saliency, and exposure features between a white-balanced image and a color-corrected image. Then, they utilized the exposure fusion algorithm to obtain the final result. However, several experiments report color shifts as a result of exposure processing.

4) **Priors**: Fattal et al. [10] estimated scene radiance and derived a transmission image using single image statistics. He et al. [11] analyzed abundant natural images and determined that most color images contain a dark channel. On the basis of this finding, they proposed a scene depth information-based dehazing algorithm. The subsequent enhanced image resulted in regional contrast stretching that can cause halos or aliasing.

In the current study, we focus on dehazing methods that use a single input image instead of multiple ones. Single-image visibility enhancement remains a challenging and ill-posed problem; the obtained brightness of each pixel depends on the radiance, scattering, attenuation, and ambient illumination of the observed scene point. In this paper, we propose a novel optical imaging model and a corresponding enhancement algorithm. First, we estimate the transmission through the robust color line to predict the ambient light $A^c$. Subsequently, we develop a rational imaging model in consideration of the positions of the camera and the imaging plane. Finally, the effects of hazing are removed by using a semi-globally adaptive filter (SAF).

3. Light Propagation Model

The atmospheric light or ambient light traveling through air is a source of illumination in the natural environment. Suppose that the intensity of light $W(x)$ after wavelength attenuation can be formulated as per the energy attenuation model as follows:

$$E_W^c(x) = E_A^c(x) + E^n(x), \quad c \in \{r,g,b\}, \tag{1}$$

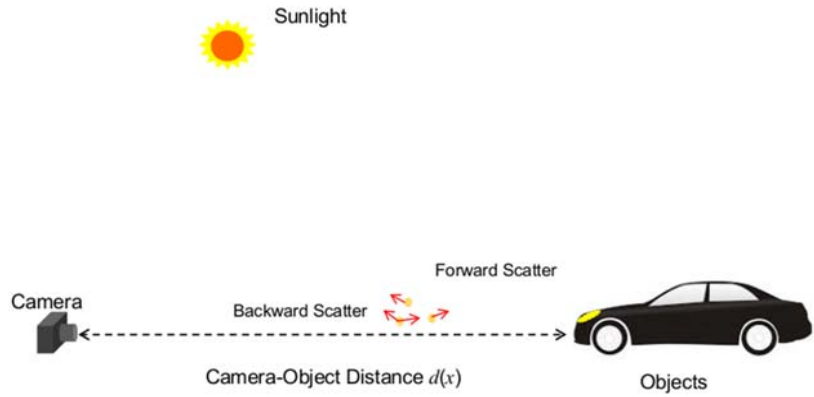

Fig. 1: Schematic of the imaging model on foggy days.

where $E_W^c(x)$ is the amount of illumination at pixel $x$ and $E_A^c(x)$ is the amount of atmospheric light illumination at pixel $x$, $c \in \{r,g,b\}$. $E^n(x)$ is the noisy illumination, such as the illumination from street light. Absorption and scattering occurs in the process of camera-to-object transmission. Supposing that the light reflection rate is $\rho^c(x)$, the total illumination of ambient light $E_W^c(x)$ is:

$$E_W^c(x) = E_A^c(x) \cdot \rho^c(x), c \in \{r,g,b\}. \tag{2}$$

As per the Koschmieder model [12], the scene image $I^c(x)$ captured by the camera can be formulated as follows:

$$I^c(x) = E_A^c(x) \cdot \rho^c(x) \cdot T^c(x) + \left(1 - T^c(x)\right) A^c, c \in \{r,g,b\}, \tag{3}$$

where the inhomogeneous background $A^c$ represents ambient light and $T^c(x)$ is the transmission disparity.

As indicated in Eq. (3), we suppose that the clean image $J^c(x)$ reflected from pixel $x$ is:

$$J^c(x) = E_A^c(x) \cdot \rho^c(x), \quad c \in \{r,g,b\}. \tag{4}$$

By substituting Eq. (4) into Eq. (3), we can obtain

$$I^c(x) = J^c(x)T^c(x) + (1 - T^c(x))A^c, c \in \{r,g,b\}. \tag{5}$$

The solution of the Koschmieder model can thus be easily obtained to solve Eq. (5). Fig. 1 shows a schematic of the proposed model. To improve image quality, we follow the processing flowchart depicted in Fig. 2.

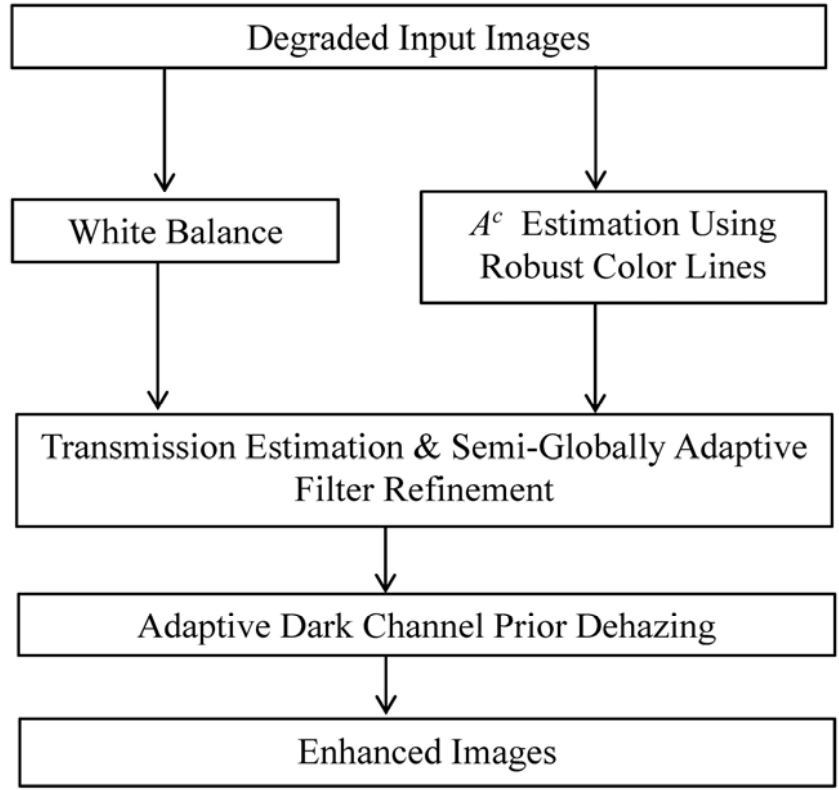

Fig. 2: Block diagram of the proposed image dehazing method.

4. Proposed Approach

### 4.1 Robust Estimation of Ambient Light

The ambient light $A^c$ in Eq. (5) is often estimated to be the brightest color in an image [9–11]. However, objects in some scenes (e.g., light-emitting diode lights) are brighter than ambient light because of artificial lighting and flash. Unfortunately, traditional approaches to predict background light may generate unfavorable results. To estimate background light reliably, Kim et al. [13] proposed a hierarchical searching method based on quad-tree subdivision. For each rectangular region, the minimization distance between (255, 255, 255) and the region is predicted. This method can identify the brightest region in the image. However, it also produces unsatisfactory results if the brightest pixels are also the objects of the image. Fattal et al. [14] presented color lines that are generated based on the global regularity approach. Nonetheless, colors shift significantly if the image reflects a single surface albedo or contains rich detail.

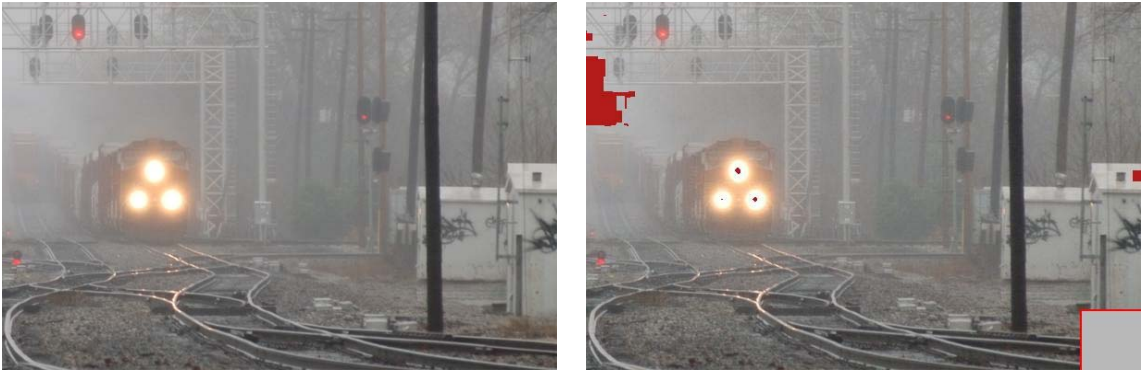

(a) Input image.          (b) Inaccurately estimated atmosphere light [17].

Fig. 3: Sample of erroneously estimated atmosphere light by the traditional dehazing model.

To address these problems, we propose an improved ambient light estimation method in this paper. First, we remove the highlight regions through applying a compensated filter to the white balanced image. The radiance value $\tilde{I}^c(x)$ of a given pixel $x$ can be modeled as

$$\tilde{I}^c(x) = E_W^c(x) \cdot R(x), \qquad (6)$$

where $E_W^c(x)$ is the amount of illumination at point $x$. $R(x)$ is the bidirectional reflectance distribution function. We assume that $R$ is a constant and is independent. The conversion of Eq. (6) to a logarithmic scale for $\tilde{I}^c(x)$ generates the following linear filter.

$$\tilde{i}^c(x) = e_W^c(x) + r(x). \qquad (7)$$

To approximate the real scene, the median filter is used to smooth the radiance. We can assume that the scene contains a practically constant irradiance and that the reflectance depends on pixel $x$.

$$i_{MED}(x) = med(\tilde{i}^c(x)) \approx e_{MED}^c + r_{MED}(x). \tag{8}$$

where $r_{MED}(x)$ stands for an approximation of the median of reflectance. The difference of the given irradiance $\tilde{i}^c(x)$ from the median radiance $i_{MED}(x)$ can be determined through

$$g' = \tilde{i}^c(x) - i_{MED}(x) = \left(e_W^c(x) - e_{MED}^c\right) + \left(r(x) - r_{MED}(x)\right). \tag{9}$$

Once Butterworth low-pass filter (LPF) is applied, Eq. (9) can be written as

$$\tilde{g}' = LPF\{g'\} = LPF\left\{\left(e_W^c(x) - e_{MED}^c\right) + \left(r(x) - r_{MED}(x)\right)\right\}. \tag{10}$$

The reflectance $r(x)$ and $r_{MED}(x)$ mainly contain high-frequency registration residues. $e_W^c(x)$ and $e_{MED}^c$ are prominent at low frequencies. Eq. (10) can be re-written as

$$\tilde{g}' = e_W^c(x) - e_{MED}^c. \tag{11}$$

where $\tilde{g}'(x)$ can represent the regions whose illumination has been corrected. Finally, the flash or highlight-corrected image $\hat{I}^c(x)$ is obtained using

$$\hat{I}^c(x) = \tilde{I}^c(x) / \exp(\tilde{g}'(x)). \tag{12}$$

Once the highlights are removed from the image, we can estimate the ambient light $A^c$ according to the color lines. As described in [14], the orientation of the atmospheric light vector $\hat{A}^c = A^c / \|A^c\|$ is calculated by exploiting the abundant small image patches in the image.

### 4.2 Scene Transmission Estimation

We assume that transmission is a constant in each patch. The transmission of the image is denoted as $\tilde{T}^c(x)$. The dark channel priors of each color channels are determined. We then define the minimum channel $I^e(x)$ for the input image $I^c(x)$ as

$$I^e(x) = \min_{c \in \{r,g,b\}} (\min_{y \in \Omega(x)} I^c(x)), \tag{13}$$

where $I^c(x)$ refers to a pixel $x$ in color channel $c \in \{r,g,b\}$ in the observed image and $\Omega$ indicates a sliding window in the image. On the basis of the dark channel prior [18], the rough transmission can be written as

$$\tilde{T}^c(x) = 1 - \kappa I^e(x), \tag{14}$$

where the scaling parameter $\kappa = 0.95$ for most scenes.

### 4.3 Refinement based on a Semi-Globally Adaptive Filter

In this section, we propose a semi-globally adaptive filter (SAF) to prevent the occurrence of gradient reversal artifacts of rough transmission. The filtering process of SAF begins under the guidance of an image, which is the white balanced image $I^B(x)$. Let $I_p^B(x)$ and $\tilde{T}_p^c(x)$ be the intensity value at pixel $p$ of the guided image and the transmission disparity image, respectively. Furthermore, let $w_k$ be the kernel window centered at pixel $k$, which must be consistent with the trilateral filter. SAF is then formulated as follows:

$$SAF(T_p^c(x)) = \frac{1}{\sum_{q \in w_k} W_{SAF_{pq}}(I_p^B(x))} \sum_{q \in w_k} W_{SAF_{pq}}(I_p^B(x))\tilde{T}_q^c(x), \tag{15}$$

where the function of the kernel weights $W_{SAF_{pq}}(I_p^B(x))$ can be written as

$$W_{SAF_{pq}}(I_p^B(x)) = \frac{1}{|w|^2} \sum_{k:(p,q) \in w_k} \left(1 + \frac{(I_p^B(x) - \mu_k)(I_q^B(x) - \mu_k)}{\sigma_k^2 + \varepsilon}\right), \tag{16}$$

where $\mu_k$ and $\sigma_k^2$ are the mean and variance of the guided image $I^B$ in local window $w_k$ and $|w|$ is the number of pixels in this window. When both $I_p^B(x)$ and $I_q^B(x)$ are on the same side of an edge concurrently, the weight assigned to pixel $q$ is large. When $I_p^B(x)$ and $I_q^B(x)$ are on different sides, a small weight is assigned to pixel $q$. Following the filtering, the refined transmission $T^c(x)$ is obtained.

Finally, once all of the desired parameters have been computed, the dehazed image is determined with

$$J^c(x) = \frac{I^c(x) - A^c}{\max(T^c(x), \varepsilon)} + A^c, \tag{17}$$

where $\varepsilon$ is chosen for numerical soundness (typically 0.0001) to avoid division by 0. Fig. 4 shows the estimated rough scene transmission and the refined transmission by the proposed semi-globally adaptive filter.

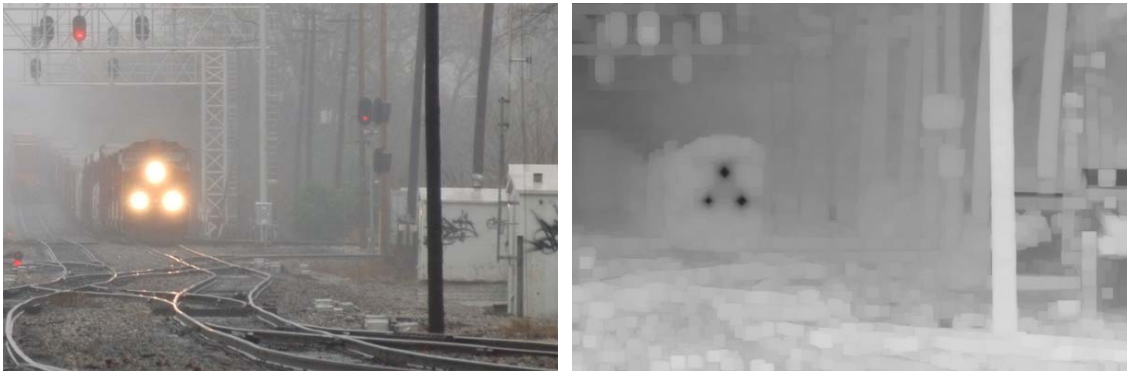

(a) Input foggy image.      (b) Estimated scene transmission.

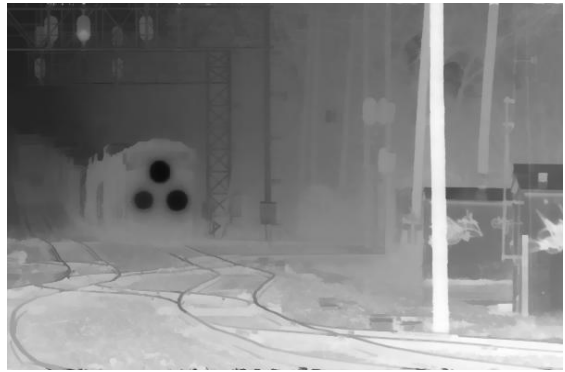

(c) Refined by Semi-Globally Adaptive Filter.

Fig.4: Transmission map estimation and refinement.

5. Experimental Results

The performance of the proposed method is evaluated both analytically and experimentally through ground-truth color patches. We also compare the proposed method with current state-of-the-art methods. The results show that the proposed method is superior to other methods in terms of haze removal and color balancing capabilities.

In the first experiment, we compare the performance of the proposed method with that of other methods to remove the scattered light from natural images. We used a Windows XP PC running on an Intel Core i7 CPU (2.0 GHz) with 4 GB RAM. The size of the images is 600 pixels × 400 pixels.

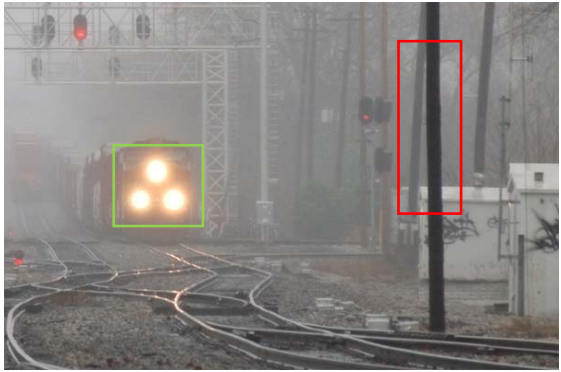

(a) Input image

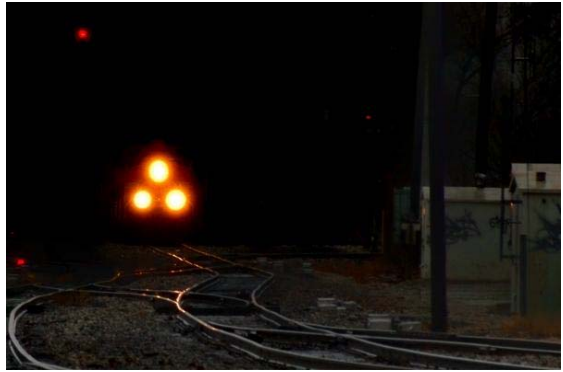

(b) He et al.'s method

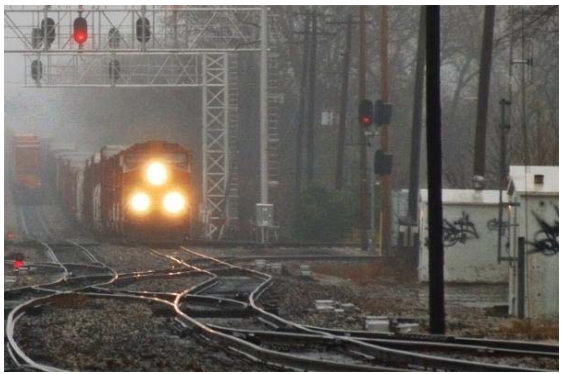

(c) Tarel et al.'s method

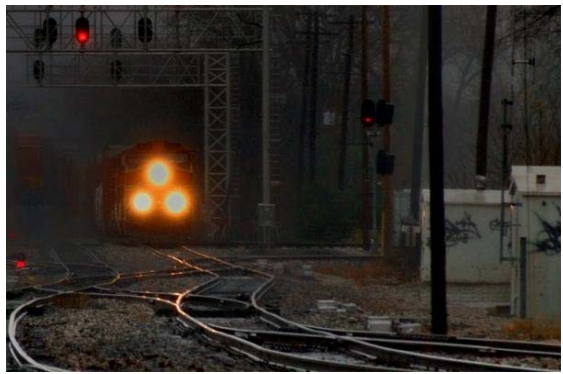

(d) Fattal et al.'s method

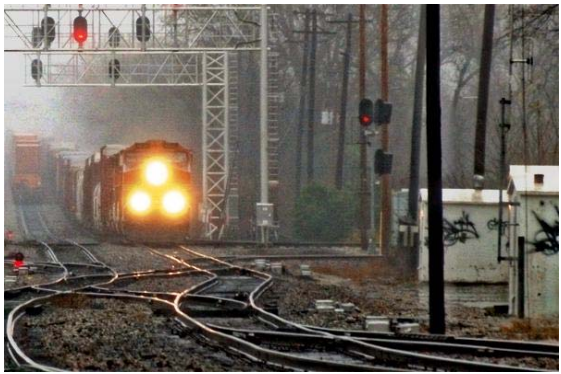

(e) Gibson et al.'s method

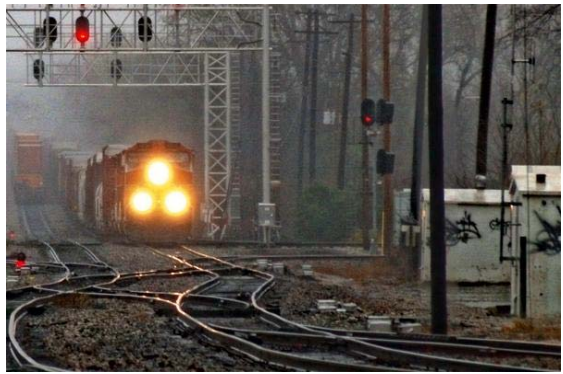

(f) Proposed method

Fig. 5: Dehazing results for *train*.

Fig. 5 shows that the images subject to the methods developed by Tarel et al. [17] and Gibson et al. [18] also remain fog. The halo effects are significant in Fig. 5(e) in particular. The contrast of the image in Fig. 5(b) (subject to He et al.'s method) is too dark because the fog levels for dehazing were overestimated. The method of Fattal et al. [14] also displays this problem because the atmospheric light was erroneously predicted. As a result, the percentage of fog or haze is inaccurately estimated. Moreover, the proposed method clarifies the object details more effectively than the other methods do. Furthermore, the methods of He et al. [11] and Tarel et al. [17] cannot estimate the transmission precisely, whereas the method of Fattal et al. over-estimates transmission. In addition, the method of Gibson et al. [18] induces transmission jumps. Fig.6 shows the cropped zoomed-in images of each methods in details. The train of He et al.'s result (Fig. 6(b)) and Fattal et al.'s result (Fig. 6(d)) is almost disappeared. Trael et al.'s method (Fig. 6(c)) conations some artifacts. The contrast of Trael et al.'s result is much lower than that of the proposed method. Gibson et al.'s result causes image blurring, and also remains some fog around the trees. Therefore, our method can remove haze more effectively than state-of-the-art methods can. The result is smooth, and the edges of the objects are clear (e.g., the lights are separated and enhanced). Halos or color shifts are not observed as well.

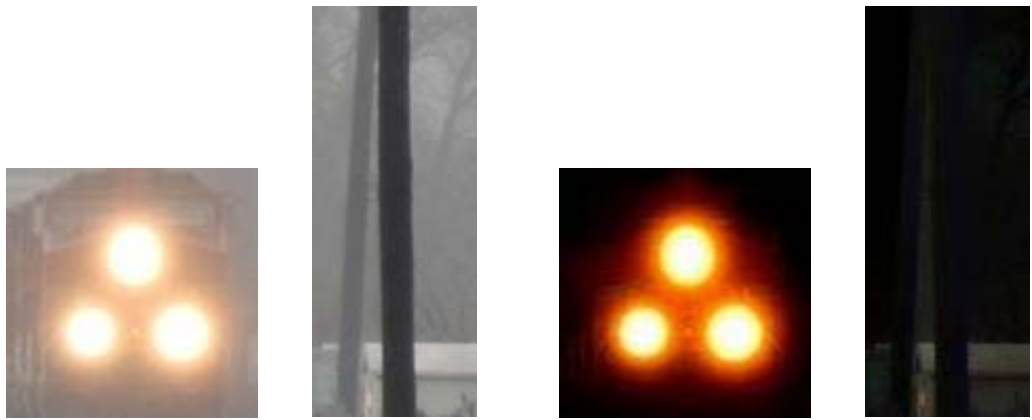

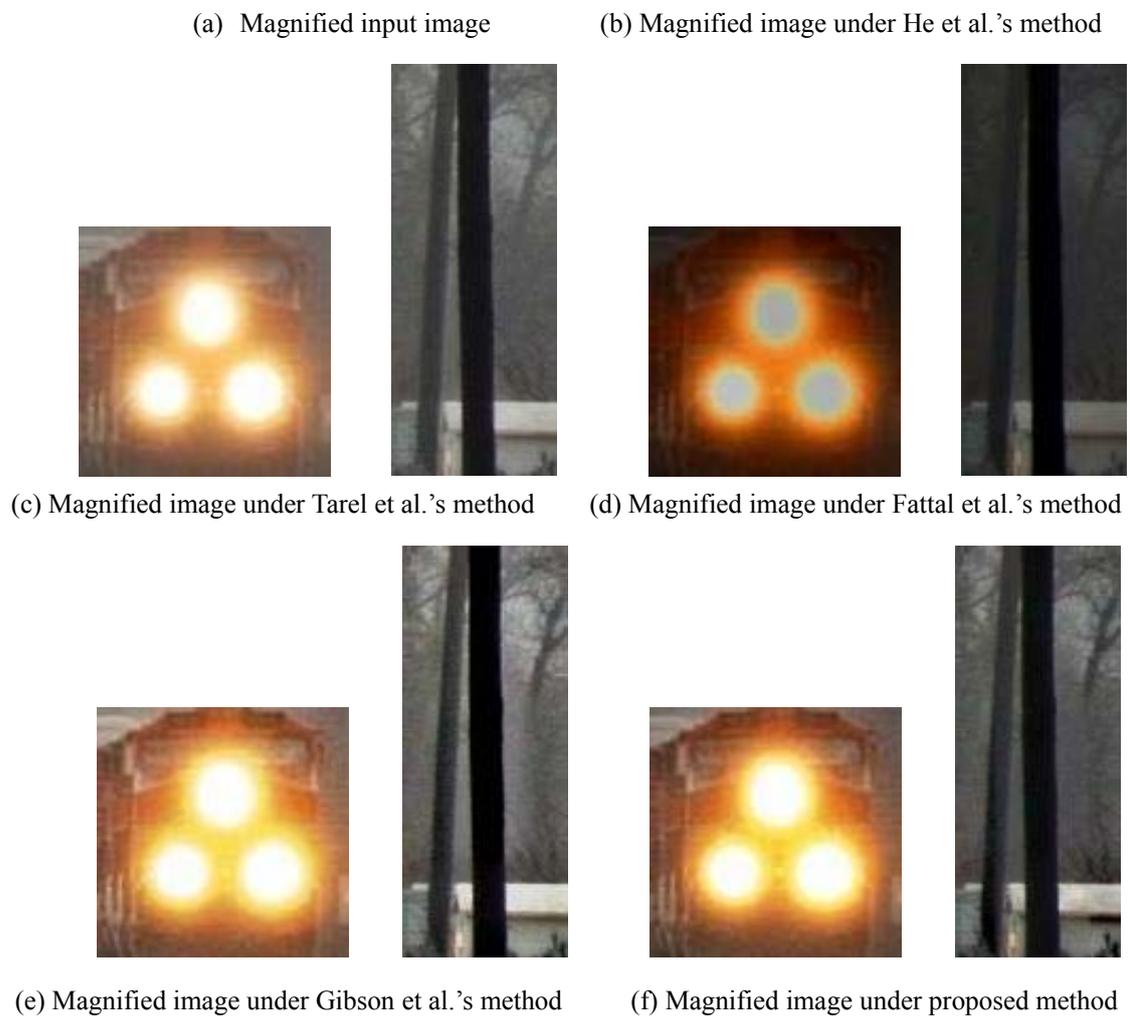

Fig. 6: Magnified images under different dehazing methods. (b) and (d) exhibit low contrast. (c) and (e) contain halos.

Fig. 7 illustrates an example to comparison of scene transmission under different methods. Compare with the state-of-the-art methods, our method estimates the transmission map much better. The edges of the objects are correctly estimated.

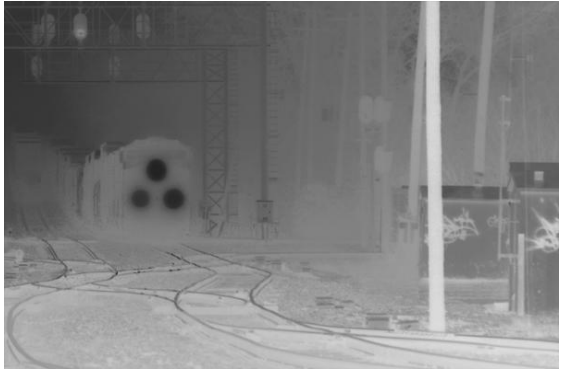 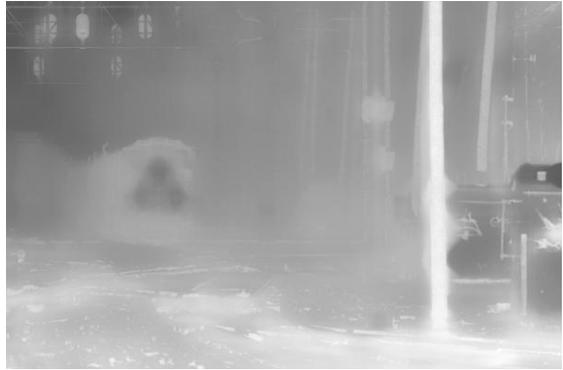

(a) He et al.'s transmission disparity.   (b) Tarel et al.'s transmission disparity.

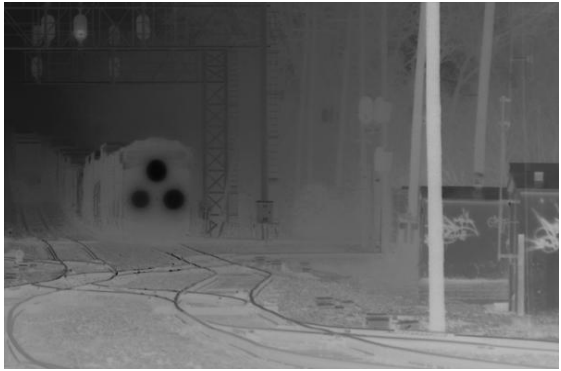 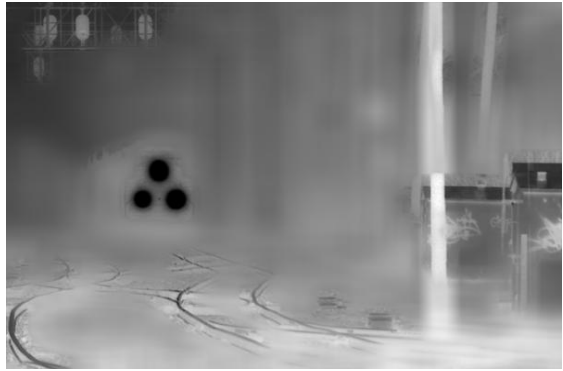

(c) Fattal et al.'s transmission disparity.   (d) Gibson et al.'s transmission disparity.

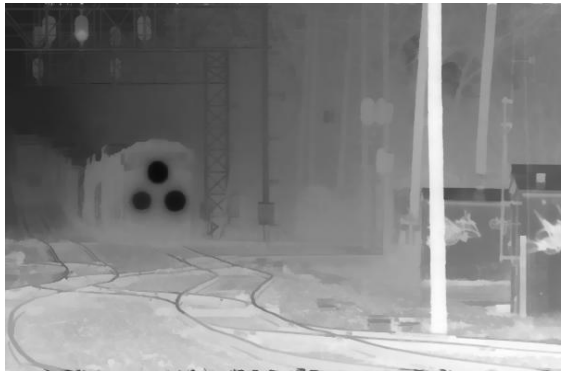

(e) Proposed transmission disparity.

Fig. 7: Transmission disparity images under different dehazing methods.

In addition to the visual analysis above, we conduct a quantitative analysis mainly from the perspective of mathematical statistics and based on the statistical parameters for the images (see

Table I). This analysis includes contrast-to-noise ratio (CNR) [19], High-Dynamic Range Visual Difference Predictor2 (HDR-VDP2) [20], new edge rate [21], and structural similarity (SSIM) [22].

CNR is a measure of image quality. It subtracts a term before determining the ratio. This measurement is important when an image displays a significant bias, such as that from haze. An image may exhibit high intensity although the features of the image are washed out as a result of haze. This image may display a high SNR metric but a low CNR metric. Therefore, CNR can measure image quality more efficiently than SNR can. The value of CNR is between 100 (best) and 0 (worst). The SSIM index determines the similarity between two images. It is a full reference metric. The value of SSIM is between 1 (best) and 0 (worst). In HDR-VDP2, the inverse Q-MOS (IQ-MOS) is equal to 100-$Q_{MOS}$. IQ-MOS value [18] is between 0 (worst) and 100 (best). We apply Nicolas et al.'s method [21] that involves computing the ratio of the gradient between the foggy image and the recovered image. This index is based on visibility level. The higher the value of the new edge rate, the better the processed result is. The results are presented in Table I.

Table I: Evaluation of the dehazed *train* image using deficient indices.

| Methods | CNR | SSIM | IQ-MOS | New Edge Rate |
| --- | --- | --- | --- | --- |
| He et al. [11] | 18.4045 | 0.1863 | 12.6911 | 33.911% |
| Trael et al. [17] | 53.2933 | **0.8561** | 47.1987 | 81.689% |
| Gibson et al. [18] | **73.9926** | 0.6707 | 70.1449 | 107.15% |
| Fattal et al. [14] | 38.4617 | 0.5128 | 55.1929 | 108.94% |
| Proposed | 62.5924 | 0.6815 | **71.8028** | **109.19%** |

Each method has its benefits. Nonetheless, the measurement of haze removal quality is a broad issue that is difficult to solve. By considering both mathematical statistics results and visual effects, we determine by a vote that our proposed method is superior to the others.

In the second experiment, we take the images of outdoor scenes from several cameras of AMOS (http://amos.cse.wustl.edu/dataset). Fig. 8 indicates that residual fog or color shifts are still observed after the dehazing operation conducted with state-of-the-art methods because of the erroneous estimation of transmission. In fact, most of the state-of-the-art methods caused the block-effects after removing the haze, which is because the error estimation of atmospheric light. In contrast, the proposed method can enhance the image better. Table II demonstrates the results.

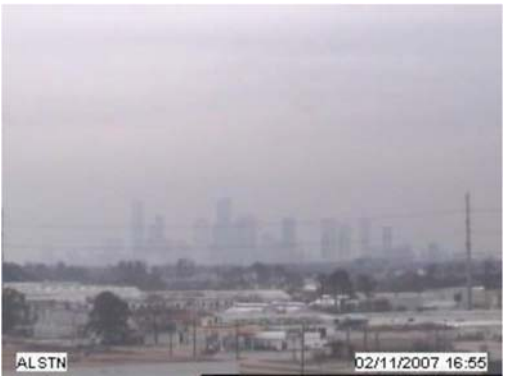
(a) Foggy image.

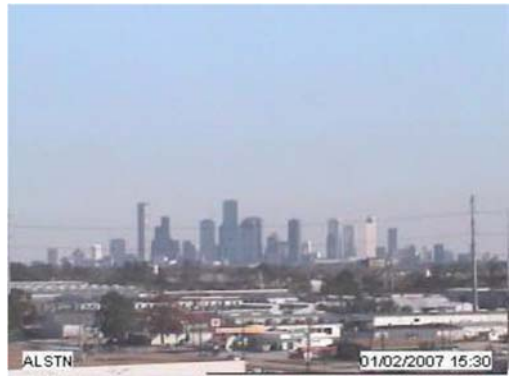
(b) Cloudy-day image.

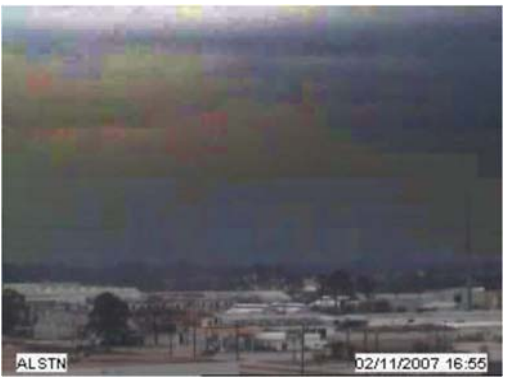
(c) He et al.'s method.

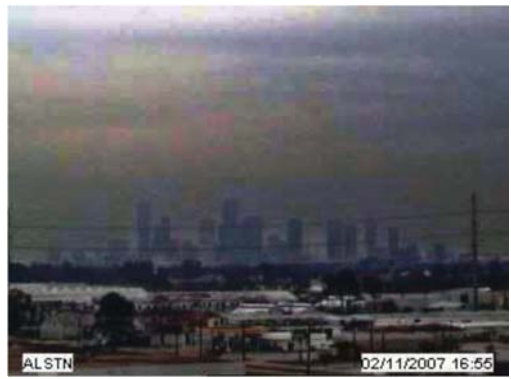
(d) Trael et al.'s method

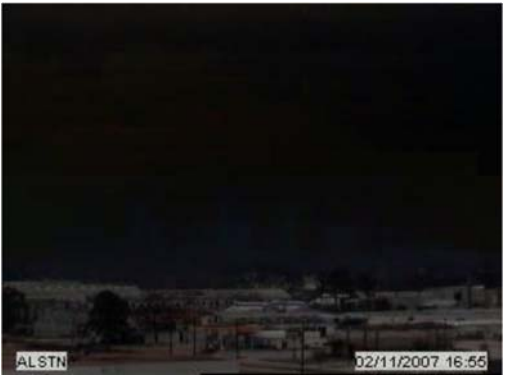
(e) Fattal et al.'s method.

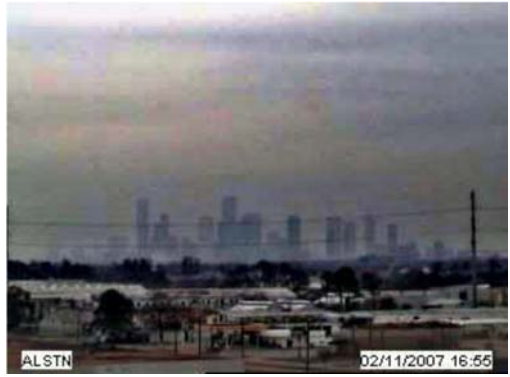
(f) Gibson et al.'s method.

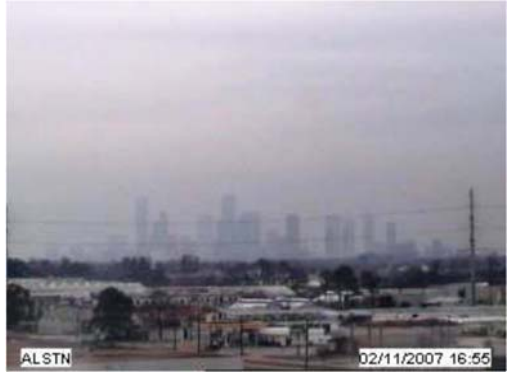
(g) Proposed method.

Fig.8: Comparison of different dehazing methods for AMOS datasets.

Table II: Evaluation of the dehazed *ALSTN* image using deficient indices.

| Methods | CNR | SSIM | PSNR | New Edge Rate |
| --- | --- | --- | --- | --- |
| He et al. [11] | 24.4239 | 0.5459 | 8.6518 | 31.961% |
| Trael et al. [17] | 34.0651 | 0.5413 | 8.6764 | **85.979%** |
| Gibson et al. [18] | **41.5051** | 0.6073 | 10.5463 | 51.416% |
| Fattal et al. [14] | 20.6612 | 0.1158 | 3.8155 | 58.633% |
| Proposed | 29.5386 | **0.7565** | **15.2440** | 56.243% |

From the above experiments, we can conclude that the proposed method outperforms the other methods for remove haze. In the third experiment, we capture images of the west door of the Tobata campus of Kyushu Institute of Technology using an in-situ camera. Fig. 9 indicates that residual fog or color shifts are still observed after the dehazing operation conducted with state-of-the-art methods because of the erroneous estimation of transmission. The proposed method can remove the fog from cloudy-day images more effectively than the other methods can.

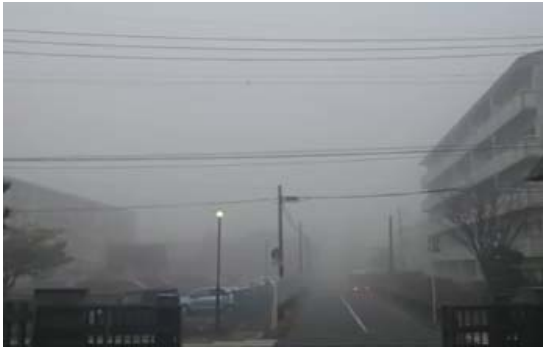 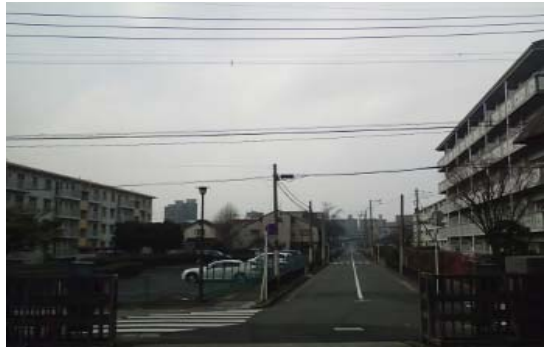

(a) Foggy image.　　　　　　　　　　　　　(b) Cloudy-day image.

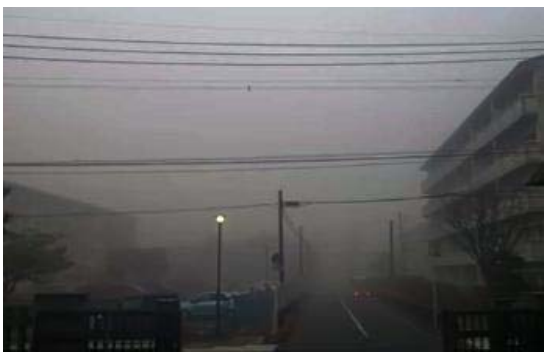 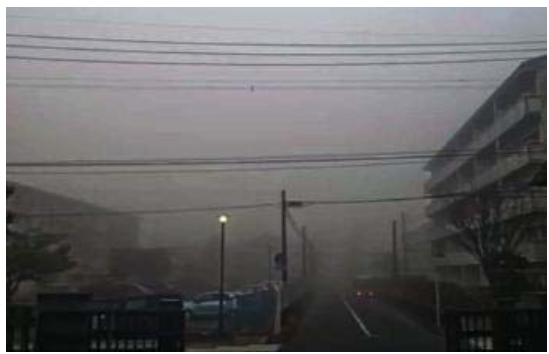

(c) He et al.'s method.　　　　　　　　　　　(d) Trael et al.'s method.

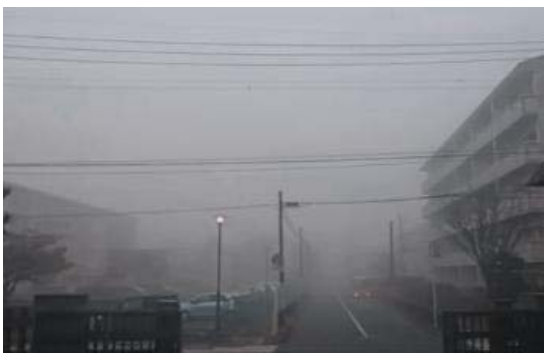 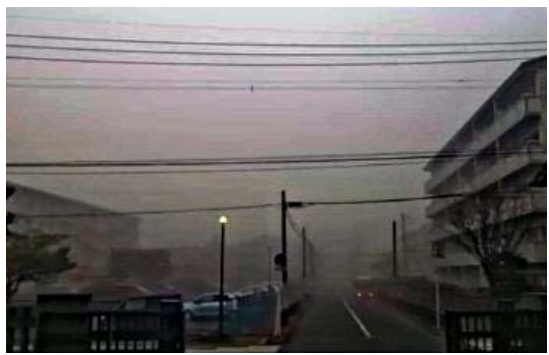

(e) Fattal et al.'s method.　　　　　　　　　(f) Gibson et al.'s method.

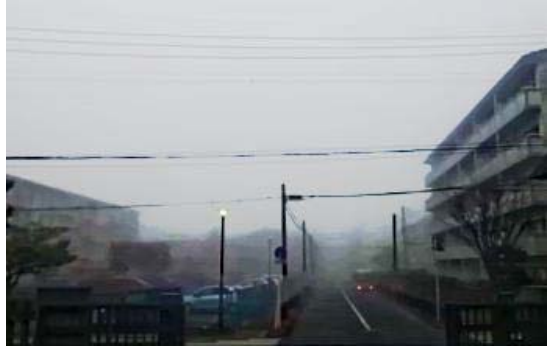

(g) Proposed method.

Fig. 9: Dehazing results for *the west door of Kyutech*.

6. Conclusion and Discussion

In this paper, we present an effective method for single image dehazing. The proposed optimized image dehazing method can generate visually compelling results. Moreover, our robust atmospheric light estimation method can produce better results than those presented in recent researches, which either assume a homogenous atmosphere or apply a difficult threshold to select the area part of the image. The proposed method can also avoid color distortion successfully. We also use a SAF algorithm to refine the transmission. The experimental results show that our algorithm is effective and is easier to compute than other guided filters are.

The proposed method has been tested on a large set of images in a natural traffic scene environment. The recovered results are visually pleasing, with no blocks or halos. In the future, we will improve the algorithm further.